\documentclass[lettersize,journal]{IEEEtran}
\usepackage{cite}
\usepackage{amsmath,amsfonts}
\usepackage{dsfont}
\usepackage{amssymb}  
\usepackage{algorithmic}
\usepackage{algorithm}
\usepackage{array}
\usepackage[caption=false,font=footnotesize,labelfont=rm,textfont=rm]{subfig}
\usepackage{textcomp}
\usepackage{stfloats}
\usepackage{url}
\usepackage{verbatim}
\usepackage{graphicx}
\usepackage{float}
\usepackage{booktabs}
\usepackage{multirow}
\usepackage[colorlinks,
linkcolor=black,
anchorcolor=black,
citecolor=black
]{hyperref}
\def\BibTeX{{\rm B\kern-.05em{\sc i\kern-.025em b}\kern-.08em
    T\kern-.1667em\lower.7ex\hbox{E}\kern-.125emX}}
\usepackage{balance}
\usepackage{etoolbox}
\makeatletter
\patchcmd{\@makecaption}
  {\scshape}
  {}
  {}
  {}
\makeatother

\begin{document}
\title{CONSS: Contrastive Learning Approach for Semi-Supervised Seismic Facies Classification}
\author{Kewen Li, Wenlong Liu, Yimin Dou, Zhifeng Xu, Hongjie Duan, Ruilin Jing 
\thanks{
The corresponding author is Kewen Li. likw@upc.edu.cn
\par
Kewen Li, Wenlong Liu, Yimin Dou, Zhifeng Xu, College of computer science and technology, China University of Petroleum (East China) Qingdao, China.
\par
Hongjie Duan, Ruilin Jing, Shengli Oilfield Company, SINOPEC Dongying, China.
\par
This work was supported by grants from the National Natural Science Foundation of China (Major Program, No.51991365),and the Natural Science Foundation of Shandong Province, China (No.ZR2021MF082).
}}

\markboth{}%
{How to Use the IEEEtran \LaTeX \ Templates}

\maketitle

\begin{abstract}
Recently, seismic facies classification based on convolutional neural networks (CNN) has garnered significant research interest. 
However, existing CNN-based supervised learning approaches necessitate massive labeled data. 
Labeling is laborious and time-consuming, particularly for 3D seismic data volumes. 
To overcome this challenge, we propose a semi-supervised method based on pixel-level contrastive learning, termed CONSS, 
which can efficiently identify seismic facies using only 1\% of the original annotations. 
Furthermore, the absence of a unified data division and standardized metrics hinders the fair comparison of various facies classification approaches. 
To this end, we develop an objective benchmark for the evaluation of semi-supervised methods, 
including self-training, consistency regularization, and the proposed CONSS. 
Our benchmark is publicly available to enable researchers to objectively compare different approaches.
Experimental results demonstrate that our approach achieves state-of-the-art performance on the F3 survey. 
Our all codes and data are available at \url{https://github.com/upcliuwenlong/CONSS}.
\end{abstract}

\section{Introduction}
\IEEEPARstart{S}{eismic} facies classification refers to the interpretation of facies type from the seismic reflector information.
It is an important first step in exploration, prospecting, reservoir characterization, and field development.
Data from core and well log offer a vertical perspective that can aid in the interpretation of seismic facies.
However, due to the expensive cost of drilling operation, direct facies information is scarce.
Alternatively, manual assignment of seismic facies based on seismic attributes is possible, 
yet remains a highly subjective process that relies heavily on the experience of the seismic interpreter.
\par
\begin{figure}[!t]
\centering
\includegraphics[width=3.5in]{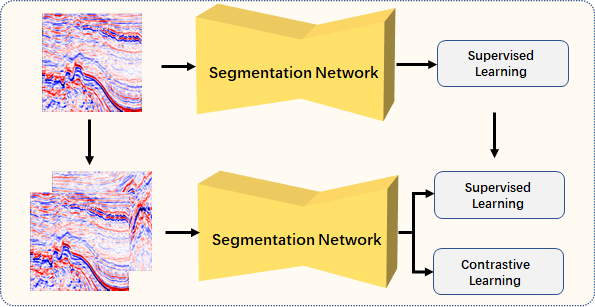}
\caption{Comparison of the fully supervised method with the proposed CONSS. Fully supervised methods utilize labeled seismic facies data. 
CONSS extends the contrastive learning branch to learning unlabeled data features.}
\label{fig_abs}
\end{figure}
Deep learning has garnered significant popularity for seismic data processing and interpretation,suah as denoising\cite{gao2021seismic}, 
inversion\cite{li2019deep}, interpolation\cite{dou2023mda} ,fault detection\cite{dou2021attention}\cite{dou2022md}, etc. 
The powerful feature extraction and representation capabilities of neural networks enable deep learning methods to mitigate human subjectivity. 
Zhao et al.\cite{zhao2018seismic} reviewed seismic facies classification and reservoir prediction approaches based on deep convolutional neural networks. 
Alaudah et al.\cite{alaudah2019machine} proposed a facies classification framework based on deconvolution network. 
These supervised deep learning methods are dependent on labeled data and are incapable of acquiring knowledge from unlabeled data. 
However, the labeling of 3D seismic data volumes is a demanding and time-intensive task, often requiring geological teams to devote hundreds of hours.
\par
In the majority of cases, there exists a scarcity of labeled data with an abundance of unlabeled data. 
Training with unlabeled data will minimize the labeling cost, 
which is a primary objective of semi-supervised learning. 
Saleem et al.\cite{saleem2019facies} implemented a semi-supervised approach based on self-training. 
They use a model trained on labeled seismic data to predict the label of unlabeled seismic data, with the predicted label subsequently added to the training set as a pseudo-label for retraining.  
Self-training is categorized as an offline learning method and is susceptible to the influence of noise pseudo-labels, leading to potential performance saturation.
Consistency regularization represents another semi-supervised paradigm, 
which operates under the foundational assumption that slight perturbations ought not to produce significant changes in the model's output. 
In our experiments, we implement the cross pseudo supervision\cite{chen2021semi} (CPS), 
which yielded superior outcomes when compared to self-training in the F3 survey. 
However, CPS necessitates the alternative training of two models, thereby resulting in a comparatively higher training overhead.
\par
Contrastive learning\cite{chen2020simple}\cite{liu2021bootstrapping}\cite{he2020momentum} is self-supervised learning in which a neural network is trained to identify the similarities and differences between different inputs. 
The main idea of contrastive learning is to learn a representation that can distinguish between positive and negative sample pairs.
In contrastive learning, a positive pair is a pair of samples that are similar to each other, 
while a negative pair is a pair of samples that are dissimilar. 
The network is trained to maximize the similarity between positive pairs and increase the dissimilarity between negative pairs. 
Kiran et al.\cite{kokilepersaud2022volumetric} proposed a pre-training method based on contrastive learning. 
Specifically, they divide the 3D seismic data volume into multiple blocks, 
and seismic data slices located within the same block were considered to be mutually positive samples, 
while slices extracted from distinct blocks were treated as negative samples. 
This method involves two stages of training, namely self-supervised pre-training and supervised fine-tuning.
\par
Different from pre-training methods based on slice-level contrastive learning, 
we propose a semi-supervised method based on fine-grained pixel-level contrastive learning, termed CONSS.
We introduce a confidence-based sampling strategy to mitigate the influence of noisy pseudo-labels in the training. 
We sample pixel-level points of different categories in regions with high classification confidence to construct positive and negative samples. 
By minimizing the contrastive loss, the network is encouraged to decrease the intra-class distance and increase the inter-class distance, 
thereby generating sharper decision boundaries to enhance the accuracy and robustness of the classification process.
Compared to self-training, our method is end-to-end without retraining.
Furthermore, unlike CPS, our approach utilizes a single network rather than two, 
simplifying the training process and reducing the overall computational burden. 
\par
It is worth mentioning that Liu et al.\cite{liu2021deep}\cite{liu2021semi} proposed a method using deep autoencoder (DAE) with classification regularization, 
aiming to minimize the ratio of the trace of intra-class and inter-class variance matrices using few labeled data. 
This approach seeks to maximize the inter-class distance and minimize the intra-class distance, 
which aligns with the objectives of CONSS. 
However, contrastive learning is fundamentally different from DAE. 
Specifically, contrastive learning focuses on directly distinguishing between positive and negative samples through individual discrimination, 
without computing sample variance matrices. 
Additionally, our confidence-based strategy enables CONSS to compute contrastive loss on unlabeled data. 
Finally, our method is end-to-end without greedy layer-wise pre-training, 
making it a more efficient and streamlined solution compared to DAE-based approaches.

\begin{figure*}[!t]
\center{\includegraphics[width=18cm] {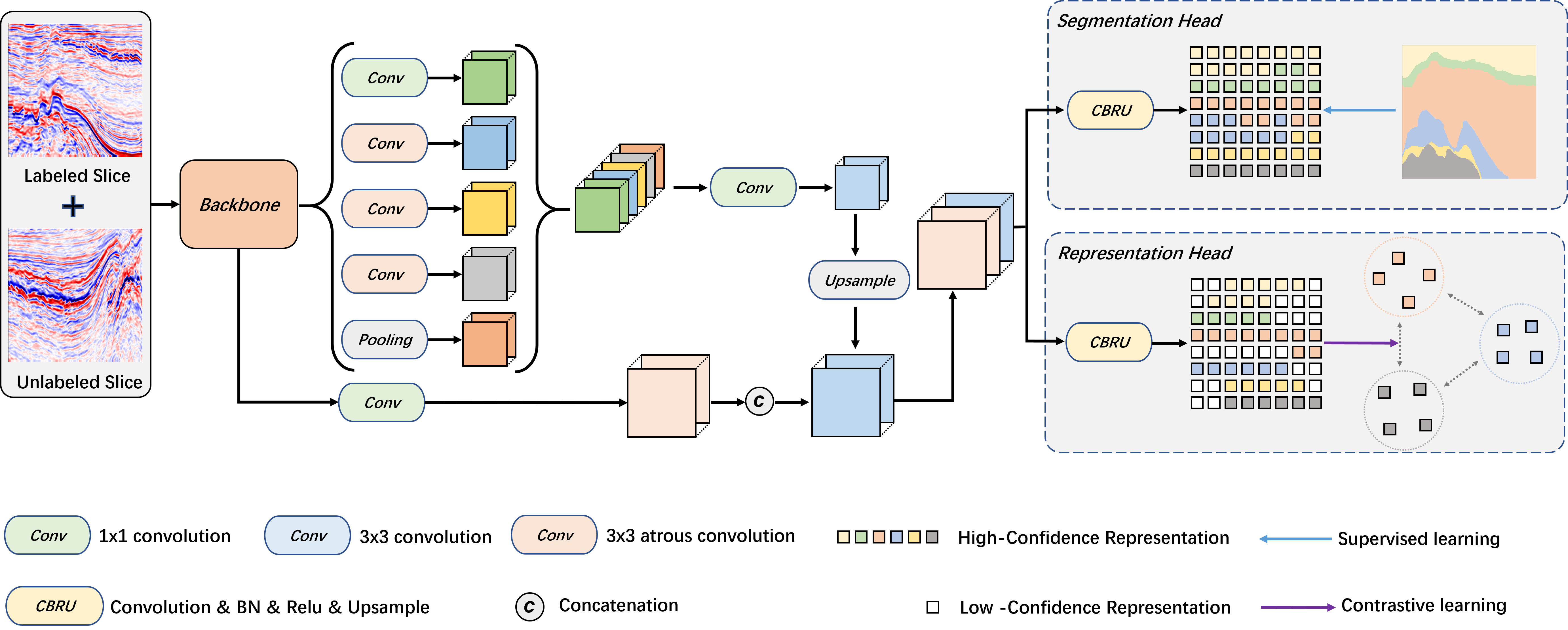}}
\caption{The CONSS pipeline consists of a backbone network that extracts features, 
a segmentation head for fully supervised learning, and a representation head for contrastive learning. 
The representation head imposes additional constraints on network learning by optimizing the feature space distribution through contrastive learning. }
\label{fig_network}
\end{figure*}

\par
In summary, the contributions of this letter are threefold:
\begin{itemize}
\item{We propose an end-to-end contrastive learning approach for semi-supervised seismic facies classification, 
which extends an additional branch that enables efficient semi-supervised learning by a plug-and-play way.}
\item{We introduce a confidence-based pixel-level sample strategy, 
which presents a promising solution for efficient negative sample generation in the presence of noisy pseudo-labels.}
\item{We develop a challenging benchmark that encompasses a unified data division and several distinct semi-supervised facies classification methods.
We have made our data and codes publicly available to facilitate fair comparisons.}
\end{itemize}

\section{Approach}
The proposed CONSS extends a branch that calculates the contrastive loss in a plug-and-play way. 
The segmentation head is responsible for delivering the predicted category and confidence score of unlabeled data to the representation head, 
which, in turn, optimizes the seismic facies representation for the segmentation head via contrastive learning. 
Notably, the two branches work in tandem to leverage each other's strengths. 
The total loss (Eq.\ref{loss}) is obtained by summing the supervised loss and the contrastive loss.
\begin{equation}
\label{loss}
\mathcal{L} = \mathcal{L}_{sup} + \mathcal{L}_{con}
\end{equation}

\subsection{Supervised Loss}
The segmentation head is formulated as a nonlinear transformation denoted by $\varphi (\mathbb{F}_{p_x,p_y}) \Rightarrow \mathbb{L}_{p_x,p_y}$. 
The objective of the mapping function $\varphi$ is to map the feature $\mathbb{F}_{p_x,p_y}$, extracted by the backbone network, to the logits $\mathbb{L}_{p_x,p_y}$,
while the position in the segmentation map $M_{seg}$ is denoted by ($p_x,p_y$).
For labeled seismic data ,
the pixel-level cross-entropy loss is calculated by computing the softmax posterior probability distribution $\mathcal{P}$  (Eq.\ref{softmax}). 
To avoid overconfidence, we use label smoothing as regularization(Eq.\ref{ce}). 
\begin{equation}
\label{softmax}
\mathcal{P}(i|\mathbb{L}_{p_x,p_y}) = \frac{e^{\mathbb{L}^i_{p_x,p_y}}}{\sum^{classes}_{i=1}{e^{\mathbb{L}^i_{p_x,p_y}}}}
\end{equation}
\par
\begin{equation}
\begin{aligned}
\label{ce}
\mathcal{L}_{sup} = &-\frac{1}{HW} \sum^H_{p_y=1} \sum^{W}_{p_x=1} \sum^{classes}_{i=1}\{\mathds{1}[i=g_t] \cdot log\mathcal{P}(i|\mathbb{L}_{p_x,p_y}) \\ 
&+ \mathds{1}[i \neq g_t]\cdot\varepsilon log\mathcal{P}(i|\mathbb{L}_{p_x,p_y})\}
\end{aligned} 
\end{equation}
where $\mathbb{L}^i_{p_x,p_y}$ represents the $i$th element of vector $\mathbb{L}_{p_x,p_y}$, 
$g_t$ represents the correct seismic facies class,
and the $\varepsilon$ is the label smoothing factor.
\subsection{High-confidence region}
For unlabeled seismic data, although the ground truth cannot be determined, 
the high-confidence region can still be regarded as reliable. 
We divide the high-confidence region into two disjoint regions, namely the weak high-confidence region and the strong high-confidence region.
\par
The selection criteria for weak high-confidence regions of class $c$ can be described as follows: 
(a) The probability of class $c$ is greater than $t_w$ but less than $t_s$. 
(b) The predicted probability of class $c$ is the highest among all classes. 
These conditions can be mathematically formulated as a ternary function (Eq.\ref{muw}) that takes into account the position $(p_x,p_y)$ and class $c$ of the pixel.
\begin{equation}
\begin{aligned}
\label{muw}
mask^w_u(p_x,p_y,c) &= \mathds{1}[ t_s > \mathcal{P}(i=c|\mathbb{L}_{p_x,p_y}) > t_w] \\
&{\;\;} \cdot \mathds{1}[ \mathop{\arg\max}\limits_{i} (\mathcal{P}(i|\mathbb{L}_{p_x,p_y})) = c]
\end{aligned}
\end{equation}
where $(p_x,p_y)$ is the position in the segmentation map.\par
\par
As for strong high confidence regions:
\begin{equation}
\begin{aligned}
\label{mus}
mask^s_u(p_x,p_y,c) &= \mathds{1}[ \mathcal{P}(i=c|\mathbb{L}_{p_x,p_y}) > t_s] \\
&{\;\;} \cdot \mathds{1}[ \mathop{\arg\max}\limits_{i} (\mathcal{P}(i|\mathbb{L}_{p_x,p_y})) = c]
\end{aligned}
\end{equation}
\par
The final weak (Eq.\ref{ruw}) or strong (Eq.\ref{rus}) high-confidence region consists of location coordinates that satisfy the selection criteria:
\begin{equation}
\label{ruw}
\mathcal{R}^w_u(c) = \{(p_x,p_y)|mask^w_u(p_x,p_y,c)=1\}
\end{equation}
\begin{equation}
\label{rus}
\mathcal{R}^s_u(c) = \{(p_x,p_y)|mask^s_u(p_x,p_y,c)=1\}
\end{equation}

\par 
As for labeled seismic data,
the ground truth is known, 
and the selection conditions are subject to an additional constraint such that the class $c$ itself is the ground truth: 
\begin{equation}
\begin{aligned}
\label{mlw}
mask^w_l(p_x,p_y,c) &= \mathds{1}[ c=g_t ]\\
&{\;\;} \cdot \mathds{1}[ \mathcal{P}(i=c|\mathbb{L}_{p_x,p_y}) < t_s ] \\
&{\;\;} \cdot \mathds{1}[ \mathop{\arg\max}\limits_{i} (\mathcal{P}(i|\mathbb{L}_{p_x,p_y})) = c]
\end{aligned}
\end{equation}
\par
\begin{equation}
\begin{aligned}
\label{mls}
mask^s_l(p_x,p_y,c) &= \mathds{1}[ c=g_t ]\\
&{\;\;} \cdot \mathds{1}[ \mathcal{P}(i=c|\mathbb{L}_{p_x,p_y}) > t_s] \\
&{\;\;} \cdot \mathds{1}[ \mathop{\arg\max}\limits_{i} (\mathcal{P}(i|\mathbb{L}_{p_x,p_y})) = c]
\end{aligned}
\end{equation}
\par
The corresponding region can be denoted as: 
\begin{equation}
\label{rlw}
\mathcal{R}^w_l(c) = \{(p_x,p_y)|mask^w_l(p_x,p_y,c)=1\}
\end{equation}
\begin{equation}
\label{rls}
\mathcal{R}^s_l(c) = \{(p_x,p_y)|mask^s_l(p_x,p_y,c)=1\}
\end{equation} 

\subsection{Positive and Negative Sample Definition}
The output of the representation head is the representation map denoted by $M_{rep}$, 
where each position is represented by a $D$-dimensional vector $\mathbb{R}_{p_x,p_y}$. 
The size of the representation map is identical to that of the segmentation map, 
i.e., both have dimensions $H$ and $W$, 
ensuring a one-to-one correspondence between the position $(p_x,p_y)$ in the segmentation map and that in the representation map.
In forward propagation, we sample $\mathcal{B}_l$ and $\mathcal{B}_u$ of the same batchsize, 
so weak (Eq.\ref{rcw}) or strong (Eq.\ref{rcs}) high-confidence positions of the class $c$ can be denoted as:
\begin{equation}
\begin{aligned}
\label{rcw} 
\mathcal{R}^w(c) = \mathcal{R}^w_u(c) \cup \mathcal{R}^w_l(c)
\end{aligned}
\end{equation}
\begin{equation}
\begin{aligned}
\label{rcs} 
\mathcal{R}^s(c) = \mathcal{R}^s_u(c) \cup \mathcal{R}^s_l(c)
\end{aligned}
\end{equation}
\par
The query vectors for each class are sampled from the weak high-confidence representations:
\begin{equation}
\label{qc}
\mathcal{Q}^c = \{\mathbb{R}_{p_x,p_y}|(p_x,p_y) \in \mathcal{R}^w(c)\}
\end{equation}
\par

For each class, only one positive sample is considered, which represents the commonality shared by all pixels of that class. Specifically, 
the positive sample for class $c$ is represented by the mean vector (Eq.\ref{ps}) of strong high-confidence representations corresponding to that class:
\begin{equation}
\label{sc}
\mathcal{S}^c = \{\mathbb{R}_{p_x,p_y}|(p_x,p_y) \in \mathcal{R}^s(c)\}
\end{equation}

\begin{equation}
\label{ps}
\mathbb{R}^{c+} = \frac{\sum\limits_{ \mathbb{R}^ \in \mathcal{S}^c} \mathbb{R}}{|\mathcal{S}^c|}
\end{equation}

\par
The strong high-confidence representations of classes other than class $c$ can be utilized as negative samples for class $c$:
\begin{equation}
\label{nss}
\mathcal{S}^{c-} = \{\mathbb{R}_{p_x,p_y}|(p_x,p_y) \in \mathcal{R}^s(i) \land i \neq c\}
\end{equation}

\subsection{Contrastive Loss}
We enforce the weak high-confidence representations of class $c$ to be in close proximity to the class center (positive sample) and distanced from its negative sample. 
To measure the similarity between two representation vectors, we utilize cosine similarity (Eq.\ref{sim}) as the metric.
\begin{equation}
\label{sim}
sim(\mathbb{R}_1,\mathbb{R}_2) = \frac{\mathbb{R}_1 \cdot \mathbb{R}_2}{|\mathbb{R}_1||\mathbb{R}_2|}
\end{equation}
The final contrastive loss is the InfoNCE \cite{oord2018representation} loss:
\begin{equation}
\begin{aligned}
\label{nce}
\mathcal{L}_{con} = &-\frac{1}{classes \cdot Q} \sum^{classes}_{c=1}\sum^Q_{i=1}\\ 
&log\frac{e^{sim(\mathbb{R}^c_i,\mathbb{R}^{c+})/\tau}}
{e^{sim(\mathbb{R}^c_i,\mathbb{R}^{c+})/\tau}+\sum^N_{n=1} e^{sim(\mathbb{R}^c_i,\mathbb{R}^{c-}_n)/\tau}}
\end{aligned}
\end{equation}
where $\mathbb{R}^c_i$ is sampled from set $\mathcal{Q}^c$, and $\mathbb{R}^{c-}_n$ is sampled from set $\mathcal{S}^{c-}$.
The coefficient $\tau$ is employed as the temperature parameter to regulate the level of discrimination against negative samples.
\section{experiment and analysis}
\subsection{Implementation Details}
The training data  was sampled from the Netherlands F3 survey. 
Seismic data slices and corresponding label slices were uniformly sampled at equal intervals along the inline direction, 
with the sampled label slices accounting for only 1\% of whole 3D volume while the remaining labels were allocated to the test set. 
Baseline refers to fully supervised training, whereas SDA\cite{olsson2021classmix} \cite{french2019semi} strong data augmentation. 
The Adam optimizer was utilized with an initial learning rate of 0.0001, 
while the batch size was set to 2. The learning rate decayed to $\frac{1}{5}$ of the previous learning rate after 4 epochs.
\subsection{Experimental Results and Comparison}
The performance of all models on the test set is reported in Table (\ref{f3_tab}). 
The improvement from self-training is extremely limited. 
The performance of CPS is relatively better, however, it requires two models to be optimized alternately. 
In contrast, our proposed CONSS method improves the score by 3.46 points and achieves the state-of-the-art performance.
\par
\begin{figure}[htbp]
\label{tes}
\begin{minipage}[t]{0.45\linewidth}
\includegraphics[width=\linewidth]{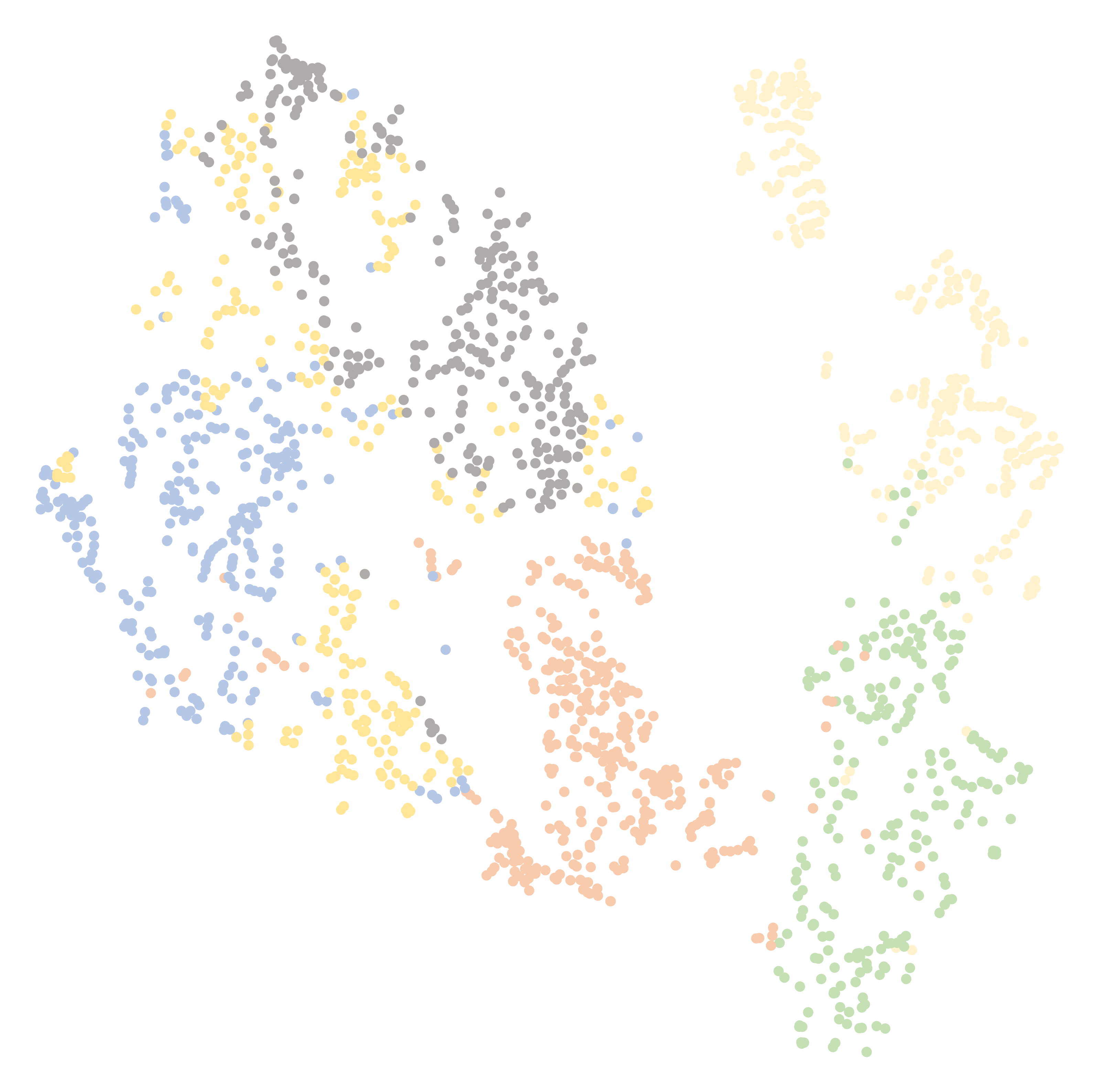} 
\end{minipage}
\hfill
\begin{minipage}[t]{0.45\linewidth}
\includegraphics[width=\linewidth]{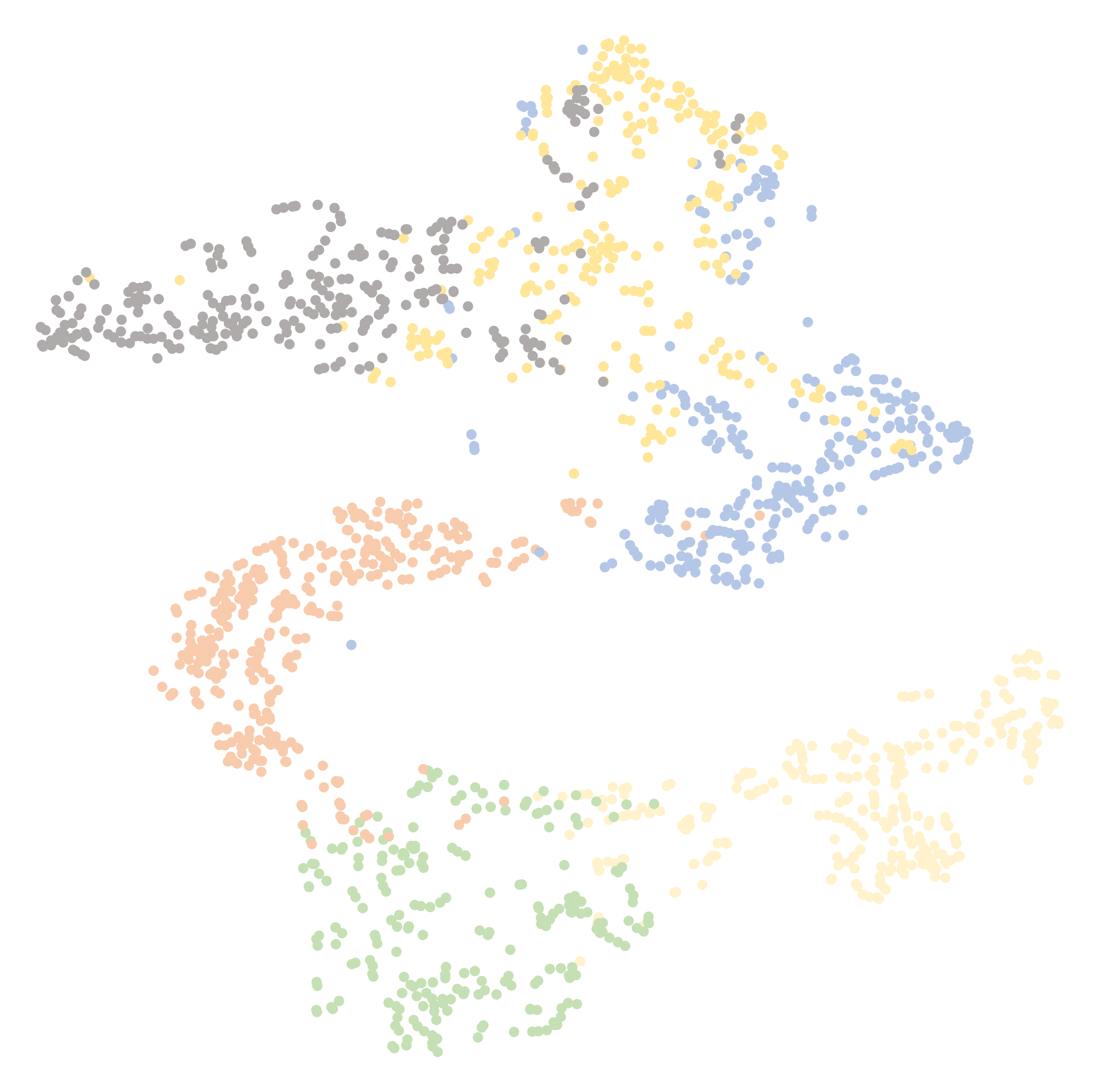}
\end{minipage}
\caption{Visualization of the feature space learned by Baseline(left) and CONSS(right), using t-SNE\cite{van2008visualizing}.} 
\end{figure}

The objective of contrastive learning is to incentivize representations belonging to the same class to approach each other and move away from other classes.
The feature space visualization (Fig.\ref{tes}) serves as an intuitive explanation for our CONSS.
\subsection{More Experiments}
The following experiments(Tab.\ref{more_f3_tab}) demenstrate that contrastive learning effectively acquires knowledge from unlabeled data, 
independently of the statistics recorded by the BN layer. 
Furthermore, the utilization of exponential moving average (EMA) in our proposed method is deemed superfluous, 
thereby simpling the training process.
\begin{table}[htpb]
\centering
\caption{}
\label{more_f3_tab} 
\begin{tabular}{cccccc}
\toprule     
labeled data & unlabeled data & $\mathcal{L}_{sup}$ & $\mathcal{L}_{con}$ & EMA & MIOU \\
\midrule
\checkmark & & \checkmark & & & 87.37 \\
\checkmark & & \checkmark & \checkmark & & 88.95 \\
\checkmark & \checkmark & \checkmark & & & 87.83\\
\checkmark & \checkmark & \checkmark & \checkmark & & {\bf90.83} \\
\checkmark & \checkmark & \checkmark & \checkmark & \checkmark & 90.21 \\
\bottomrule    
\end{tabular}
\end{table}

\begin{table*}[htbp]
\renewcommand\arraystretch{1.2}
\caption{Metrics of Netherlands F3 survey.}
\label{f3_tab}
\centering
\begin{tabular}{lccccccccccc}
\toprule     
\multirow{2}{*}{Method} & \multicolumn{6}{c}{Class Accuracy} \\
\cline{3-8}
& PA & Upper & Middle & Low & R/C & Scruff & Zechstein & MCA & FWIOU & MIOU & F1 \\
\midrule
supervised (baseline) & 97.00 & 98.72 & 94.14 & 98.93 & 91.37 & 91.40 & 77.43 & 92.00 & 94.30 & 87.37 & 92.99 \\
supervised w/ SDA & 97.06 & 98.72 & 94.52 & 98.91 & 90.38 & 91.96 & 80.22 & 92.45 & 94.42 & 87.69(+0.32) & 93.21 \\
self-training & 97.04 & 98..81 & 94.12 & 98.94 & 91.48 & 91.50 & 77.88 & 92.12 & 94.39 & 87.56(+0.19) & 93.11 \\
CPS & 97.32 & 99.02 & 93.63 & 99.15 & 93.50 & 89.12 & 88.25 & 93.78 & 94.91 & 88.77(+1.40) & 93.88 \\
CONSS & 96.45 & 99.17 & 88.09 & 99.40 & 92.93 & 88.54 & 69.43 & 89.59 & 93.33 & 84.54(-2.83) & 91.15 \\
CONSS w/ SDA & 97.61 & 99.15 & 93.32 & 98.98 & 93.96 & 93.42 & 92.70 & 95.25 & 95.43 & {\bf90.83(+3.46)} & 95.11 \\
\bottomrule     
\end{tabular}
\end{table*}

\begin{figure*}[htpb]
\centering
\captionsetup[subfloat]{labelsep=none,format=plain,labelformat=empty}
\subfloat[(a)]{\includegraphics[width=0.9in]{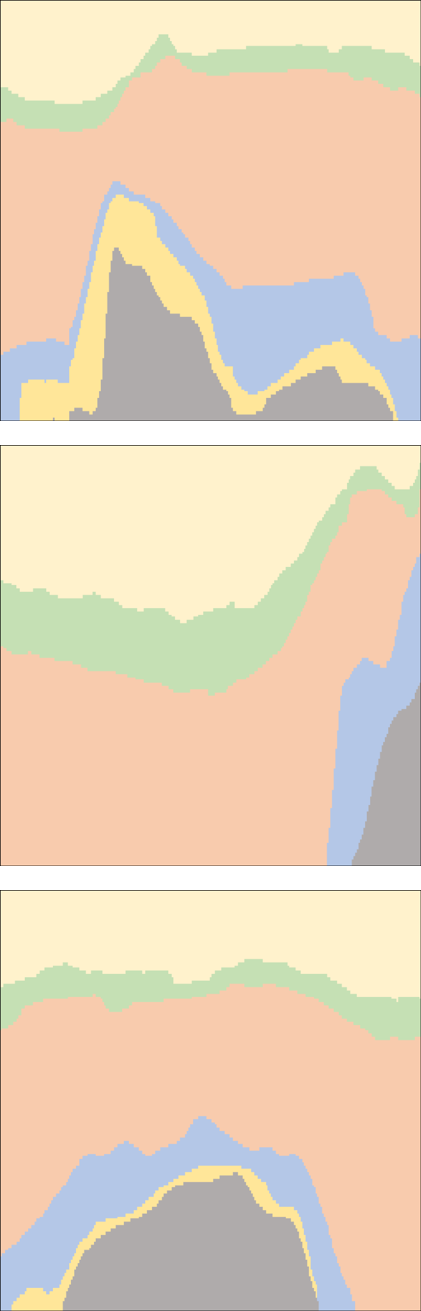}
    \label{f3_gt}}
\hspace{-0.1cm}
\subfloat[(b)]{\includegraphics[width=0.9in]{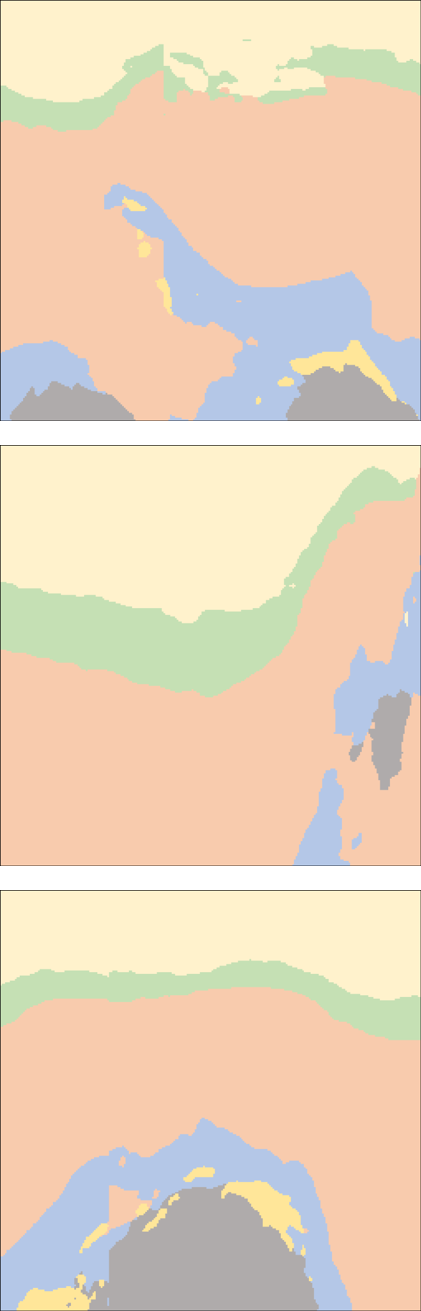}
    \label{f3_bl}}
\hspace{-0.1cm}
\subfloat[(c)]{\includegraphics[width=0.9in]{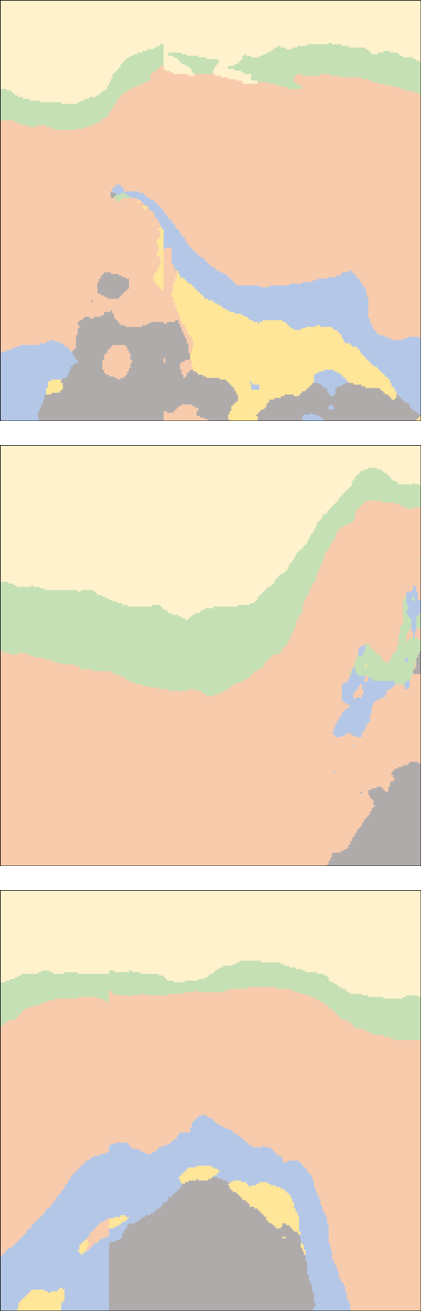}
    \label{f3_bl_sda}}
\hspace{-0.1cm}
\subfloat[(d)]{\includegraphics[width=0.9in]{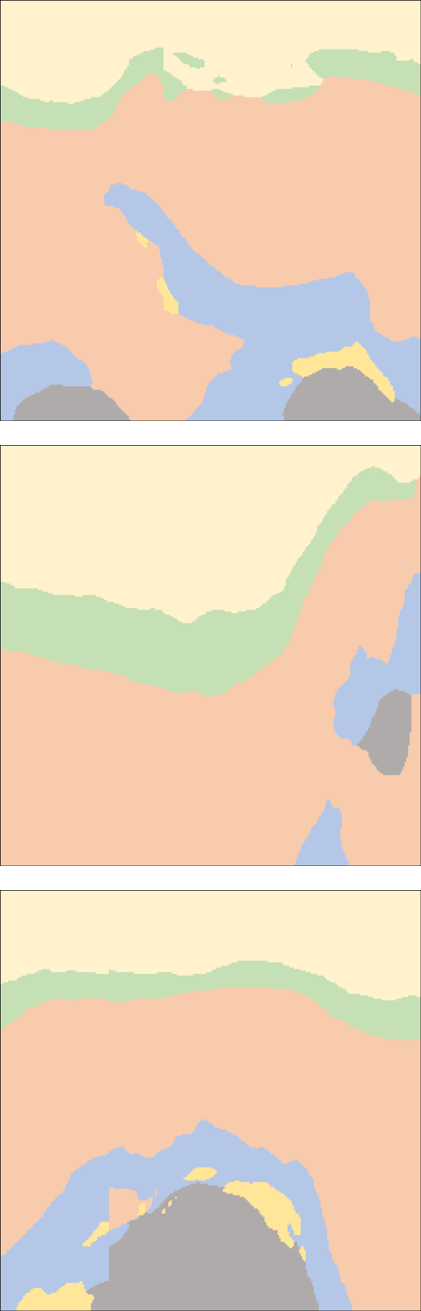}
    \label{f3_st}}
\hspace{-0.1cm}
\subfloat[(e)]{\includegraphics[width=0.9in]{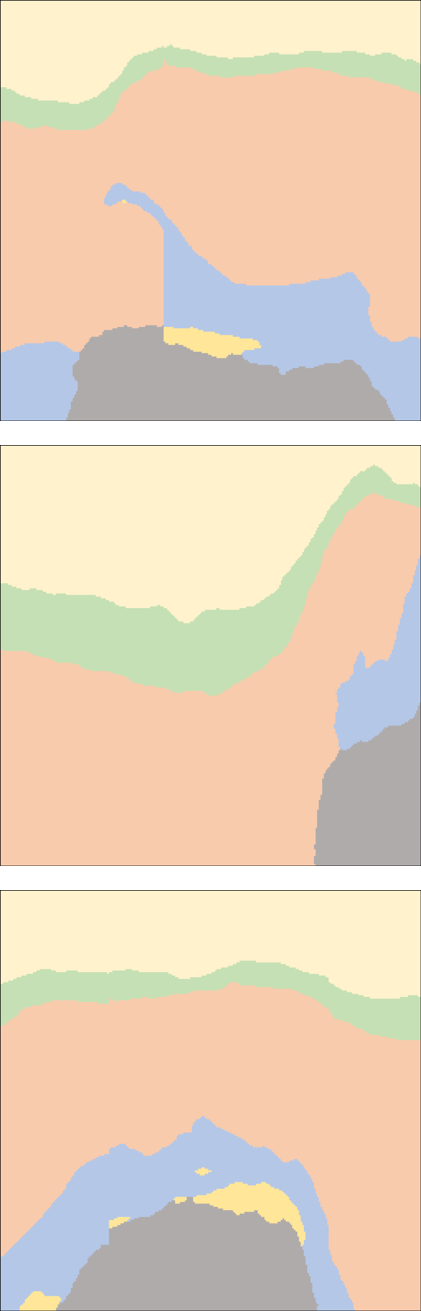}
    \label{f3_cps}}
\hspace{-0.1cm}
\subfloat[(f)]{\includegraphics[width=0.9in]{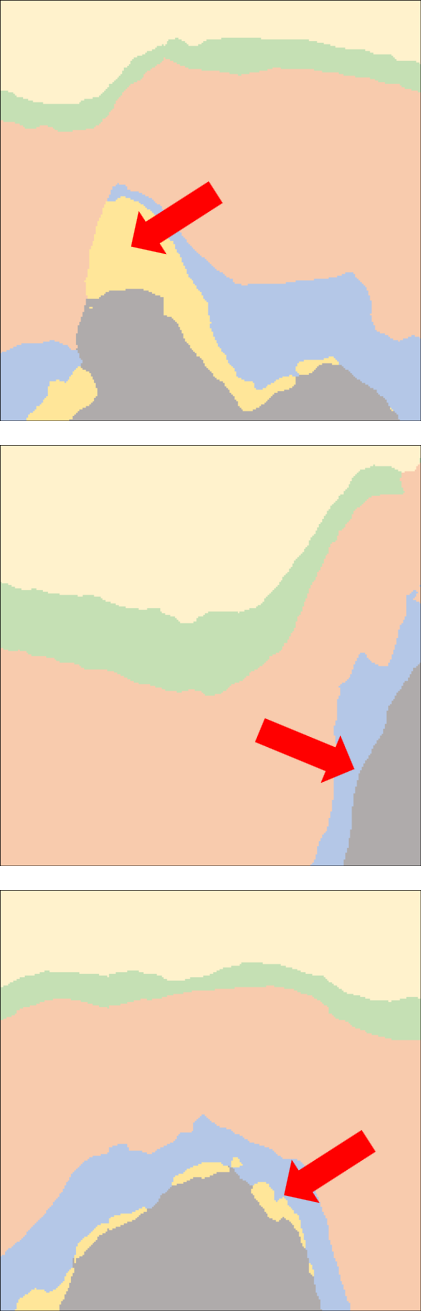}
    \label{f3_conss}}
\caption{Visualized results of different seismic facies classification methods on the Netherlands F3 survey , 
and the red arrows indicate the improvement. (a) Ground Truth. 
(b) Result by supervised. (c) Result by supervised /w SDA. (d) Result by self-training. 
(e) Result by CPS. (f) Result by CONSS /w SDA.}
\label{f3_img}
\end{figure*}

\subsection{Discussion}
We found that SDA is necessary for our CONSS.
SDA\cite{olsson2021classmix} \cite{french2019semi} is a widely adopted technique in semi-supervised learning, 
which facilitates the network to learn fundamental semantic features in perturbed images.
In this respect, its function is similar to the perturbation added to the consistency regularization.
Moreover, seismic facies are strongly associated with the depth, 
the neural network may only rely on the depth information to achieve low loss. 
By properly disrupting the strong correlation between seismic facies and depth using SDA, 
we increase the learning difficulty of model, thereby avoiding shortcut solution.
\par
It is noteworthy that the effectiveness of contrastive learning is contingent on a configuration of various factors, 
which mandates the adoption of an appropriate sample strategy, 
coupled with critical components such as SDA and label smoothing.  
Furthermore, as the number of components and hyperparameters increase, 
the complexity of the training process intensifies.
The current method is only applicable when multiple facies appear in the same slice at the same time. 
In extreme cases, when there is only one kind of facies in a slice, 
the contrastive loss cannot be calculated because there is no memory bank to store other facies.

\section{conclusion}
In this letter, we propose a contrastive learning approach for semi-supervised seismic facies classification, termed CONSS.
Our approach outperforms existing semi-supervised methods in two aspects.
Firstly, in contrast to slice-level contrastive learning approaches, 
the CONSS operates at the pixel-level, resulting in greater conservation of memory resources. 
And we introduce a fine-grained confidence-based sample strategy, which can mitigate the interference of noisy pseudo-labels.  
Secondly, unlike the self-training and CPS approaches, our method eliminates the need for retraining and alternate optimization, 
making it an end-to-end training method. 
Our proposed CONSS method achieves state-of-the-art performance with faster training efficiency. 
Additionally, we develop a open-source benchmark that will enable fair comparisons of various semi-supervised seismic facies classification methods.
\bibliographystyle{IEEEtran}
\bibliography{ref.bib}
\end{document}